\pdfoutput=1

\documentclass[11pt]{article}

\usepackage[final]{acl}

\usepackage{times}
\usepackage{latexsym}
\usepackage{amsmath}
\usepackage{amssymb}
\usepackage{tabularx}
\usepackage{multirow}
\usepackage{booktabs}
\hypersetup{urlcolor=black}

\usepackage[T1]{fontenc}

\usepackage[utf8]{inputenc}

\usepackage{microtype}

\usepackage{inconsolata}

\usepackage{graphicx}

\title{Retrieval over Classification: Integrating Relation Semantics for
Multimodal Relation Extraction}

\author{
 \textbf{Lei Hei\thanks{Equal contribution.}},
 \textbf{Tingjing Liao\footnotemark[1]},
 \textbf{Yingxin Pei},
 \textbf{Yiyang Qi},
 \textbf{Jiaqi Wang},
 \textbf{Ruiting Li},
 \textbf{Feiliang Ren\thanks{Corresponding author.}} \\
 \text{School of Computer Science and Engineering,} \\ 
 \text{Northeastern University, Shenyang 110819, China} \\
 \href{mailto:renfeiliang@cse.neu.edu.cn}{\texttt{renfeiliang@cse.neu.edu.cn}}
}

\begin{document}
\maketitle

\begin{abstract}
Relation extraction (RE) aims to identify semantic relations between entities in unstructured text. Although recent work extends traditional RE to multimodal scenarios, most approaches still adopt classification-based paradigms with fused multimodal features, representing relations as discrete labels. This paradigm has two significant limitations: (1) it overlooks structural constraints like entity types and positional cues, and (2) it lacks semantic expressiveness for fine-grained relation understanding. We propose \underline{R}etrieval \underline{O}ver \underline{C}lassification (ROC), a novel framework that reformulates multimodal RE as a retrieval task driven by relation semantics. ROC integrates entity type and positional information through a multimodal encoder, expands relation labels into natural language descriptions using a large language model, and aligns entity-relation pairs via semantic similarity-based contrastive learning. Experiments show that our method achieves state-of-the-art performance on the benchmark datasets MNRE and MORE and exhibits stronger robustness and interpretability.
\end{abstract}

\section{Introduction}

Relation extraction (RE) is a fundamental task in information extraction, aiming to identify semantic relations between entities from unstructured texts automatically\cite{re_text_1, re_text_2}. It provides crucial structured data for downstream applications such as knowledge graph construction and question answering\cite{luo2018knowledge, li2019entity, yu2020relationship}. 

However, traditional text-only RE methods face two key challenges. First, the inherent ambiguity of natural language often leads to incorrect predictions due to insufficient contextual information. Second, real-world data is frequently accompanied by visual information, and accurate relation inference in such scenarios usually requires joint reasoning over both textual and visual modalities\cite{MNRE}. These limitations motivate research in multimodal relation extraction (MRE).

\begin{figure}
    \centering
    \includegraphics[width=0.99\linewidth]{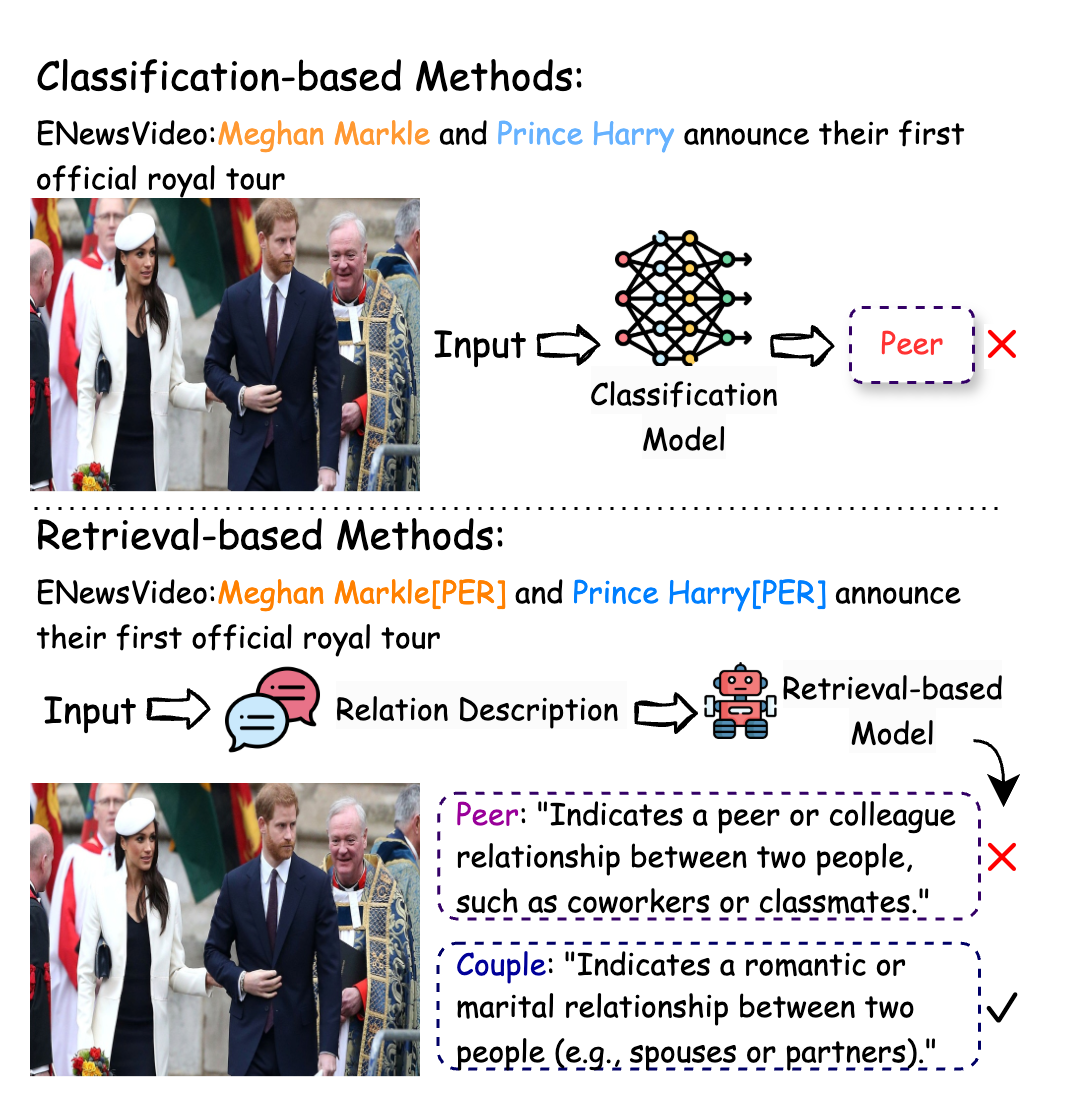}
    \caption{Comparison of classification-based and retrieval-based methods.}
    \label{fig:head_fig}
\end{figure}

Existing MRE approaches primarily follow two paradigms: modality fusion\cite{chen2022good, TSVFN, liu2024multimodal, MMIB} and modality alignment\cite{MEGA, MREISE, PROMU, VMHAN}. Despite their technical differences, both paradigms ultimately rely on mapping multimodal features into a discrete set of predefined relation categories—a classification-based framework. This paradigm faces two significant limitations.

First, it neglects structural constraints such as entity types and positions. For instance, in a location\_at relation, the subject is typically a location or an organization, while the object is usually a location. Without modeling such priors, the model must search for relations across many irrelevant entity pairs, significantly increasing reasoning difficulty and reducing classification accuracy.

Second, fixed label indices limit the model's semantic expressiveness for fine-grained relation understanding. Figure~\ref{fig:head_fig} illustrates a concrete example: due to the semantic similarity between "Peer" and "Couple" in the representation space, a classification model tends to misclassify a married couple as "Peer". This mistake stems from the limited expressiveness of discrete labels, which fail to capture semantically similar but fundamentally different relations, weakening the model's ability to distinguish subtle relational nuances.

To address these limitations, we propose a novel framework: Retrieval Over Classification (ROC), which reformulates multimodal relation extraction as a semantically-driven retrieval task.

To address the first challenge, we incorporate entity types and positions to constrain the candidate relation space. We use the Stanford NER toolkit\cite{NER_tool} to identify entity types and embed them into the input as explicit semantic prompts, guiding the model to learn type-aware semantics. Entity positions help localize entity pairs and construct centered representations. We design a type-aware multimodal encoder to jointly encode these structural cues, effectively narrowing the candidate space and improving classification accuracy.

To address the second challenge, we replace discrete relation labels with natural language descriptions to enhance semantic expressiveness. Using GPT-4o\cite{gpt4o}, we generate descriptive sentences for each relation, followed by manual verification to ensure quality. A relational semantic encoder then transforms these descriptions into global semantic representations. Compared to fixed labels, this approach enables finer-grained relation modeling and improves the model's ability to distinguish semantically similar relations.

To align entity pairs with relation semantics, we introduce a contrastive retrieval strategy. The multimodal entity pair encoder and the relation semantic encoder project their features into a shared semantic space. The model is trained to maximize the similarity between matched pairs while suppressing irrelevant candidates, enabling accurate relation prediction. This retrieval-based paradigm integrates multimodal information, maintains semantic interpretability through natural language, and mitigates the label rigidity often observed in classification-based approaches.

Our contributions are summarized as follows:

\begin{itemize}
    \item We propose ROC, reformulating multimodal relation extraction as a semantic retrieval task, offering a paradigm shift from traditional classification-based methods.

    \item We design a type-aware multimodal encoder incorporating entity type and position to effectively constrain the candidate relation space. Additionally, we construct natural language relation representations and introduce a relation semantic encoder to enhance fine-grained semantic modeling. A cross-modal contrastive retrieval mechanism aligns entity pairs with relation semantics in a shared space, enabling semantically consistent relation prediction.

    \item ROC achieves state-of-the-art performance on both the multimodal relation extraction benchmark MNRE\cite{MNRE} and the cross-modal dataset MORE\cite{MORE}, demonstrating the effectiveness and generalizability of our approach.
    
\end{itemize}

\section{Related Work}

\textbf{Modality Fusion Paradigms} Modality fusion methods aim to enable interactive learning of visual and textual features through deep neural networks. Representative approaches include HVPNeT~\cite{HVPNeT}, which introduces multi-scale visual features and leverages a dynamic gating mechanism to guide the language model in capturing image context. MMIB~\cite{MMIB}, built on variational autoencoders, incorporates mutual information maximization~\cite{MI} and the information bottleneck principle~\cite{IB} to alleviate representational discrepancies across modalities. While these methods effectively enhance cross-modal semantic perception, they fundamentally rely on mapping fused features to a predefined label space using discrete relation classifiers. This paradigm overlooks structural priors between entity types, making it challenging to model semantically continuous relationships, thus exhibiting a strong dependence on rigid labels.

\noindent
\textbf{Modality Alignment Paradigms} Modality alignment methods introduce structured intermediate representations to guide semantic mapping across modalities, thereby enhancing relation modeling. For example, MEGA~\cite{MEGA} aligns syntactic dependency structures from text with visual scene graphs~\cite{sgg_for_mega} from images to improve entity-relation extraction accuracy. MREISE~\cite{MREISE} leverages CLIP~\cite{CLIP} to construct cross-modal graph structures and optimizes their representation via an information bottleneck mechanism. These methods improve model interpretability through explicit structural constraints, but their ability to model relations remains limited by the expressiveness and applicability of the prior structures. This limitation becomes more pronounced in diverse and complex contexts, where capturing the semantic dependencies between entity types and flexibly modeling structural constraints are challenging.

\noindent
\textbf{Emerging Paradigms} In recent years, some studies have sought to overcome the limitations of traditional classification paradigms. EEGA~\cite{JMERE} introduces an end-to-end framework for joint extraction of entities and relations, removing the dependence on predefined entity labels. MOREformer~\cite{MORE} explores cross-modal object-entity relation modeling, offering greater adaptability in multimodal scenarios. However, these approaches still follow a two-stage pipeline of “feature learning followed by discrete classification,” and have yet to break through the representational bottleneck of classification-based semantic modeling fundamentally.

The above methods primarily focus on classification, emphasizing the extraction of textual and visual features while neglecting the semantic modeling of relationships themselves. Therefore, we propose a retrieval-based paradigm incorporating relational semantics which enables the model to extract multimodal features while constraining the search space for entity pairs through type constraints. It also leverages natural language descriptions to provide fine-grained semantic information, more effectively facilitating the model's learning process.

\section{Methodology}
\begin{figure*}[t]
    \centering
    \includegraphics[width=\textwidth]{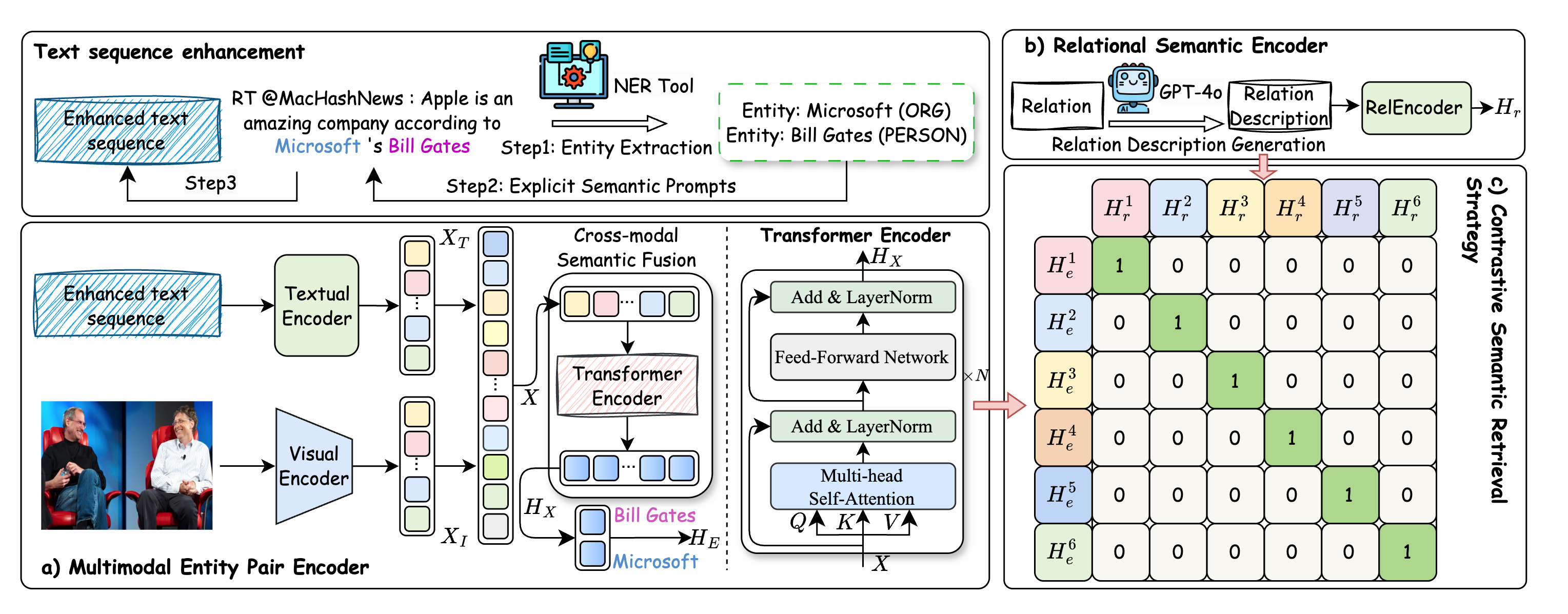}
    \caption{A multimodal relational extraction model based on relational semantic retrieval paradigm.}
    \label{fig:roc}
\end{figure*}
\subsection{Task Definition}

The task of MRE can be formally defined as follows: Given an input text sequence $T = [w_1, w_2, \dots, w_n]$ and its associated image $I$, the goal is to predict a set of relational triples $Y = \{(s, r, o)_c\}_{c=1}^C$. Here, $s \in E$ and $o \in E$ denote the subject and object entities, where $E$ is the set of all entities in the input. The relation $r \in R$ is selected from a predefined relation set $R$, and $(s, r, o)_c$ represents the $c$-th predicted relation triple.

Unlike traditional classification paradigms, we do not directly select a relation type $r$ from the discrete relation label space $R$. Instead, each relation type is represented by a \textit{natural language description}, which is encoded into a \textit{shared semantic space} via a \textit{relation semantic encoder}. Based on the fully integrated multimodal representation of the entity pair $(s, o)$, the model retrieves the relation description that is most semantically aligned in this space to determine the relation type $r$. This reformulates multimodal relation extraction as a semantics-driven multimodal retrieval task.

\subsection{Overview}

The overall architecture of our proposed ROC model is illustrated in Figure~\ref{fig:roc}. It mainly consists of three core components: (1) \textbf{Multimodal Entity Pair Encoder}: It integrates explicit entity type annotations with a Transformer-based interaction mechanism to fuse textual and visual features, enhancing the semantic representation of entities in cross-modal contexts (Section~\ref{sec:pair-encoder}); (2) \textbf{Relation Semantic Encoder}: Relation types are represented in natural language and encoded into a unified semantic space using an independent language model, which explicitly models the semantic differences between relation categories (Section~\ref{sec:rel-encoder}); (3) \textbf{Contrastive Semantic Retrieval Strategy}: A matching mechanism is established between multimodal entity pair representations and relation semantics. The model achieves more discriminative relation extraction by optimizing the semantic similarity between each entity pair and its corresponding relation description (Section~\ref{sec:retrieval}).

\subsection{Multimodal Entity Pair Encoder}
\label{sec:pair-encoder}

To constrain the candidate relation space and improve the accuracy of relation extraction, we design a type-aware multimodal entity pair encoder. It jointly models entity types and positional information to guide the model in filtering out semantically or spatially inconsistent entity combinations, while effectively integrating textual and visual features.

We use the Stanford NER tool\cite{stanford} to identify entity types in the input text. The recognized type information is embedded into the original text sequence as an explicit semantic prompt, guiding the model to perceive type priors during encoding. A pretrained BERT\cite{BERT} model encodes the enhanced textual sequence to obtain the textual feature representation $X_T$.

Meanwhile, we adopt a pretrained Vision Transformer (ViT)\cite{ViT} to extract visual features $X_I$. Then, we concatenate the textual and visual features along the sequence dimension to form a unified multimodal representation $X = [X_T; X_I]$, which is passed through multiple Transformer encoder layers. A multi-head self-attention mechanism is applied to achieve deep cross-modal semantic fusion:

\begin{equation}
H_X = \text{Encoder}(X W_Q, X W_K, X W_V)_N
\end{equation}

Where $W_Q, W_K, W_V$ are the learnable projection matrices for queries, keys, and values in the self-attention mechanism.

Let the indices of the subject and object entities in the concatenated sequence be $\tilde{s}$ and $\tilde{o}$. Based on their positions in the text, we extract the contextualized representations $(H_s, H_o)$ of the subject entity $s$ and the object entity $o$ from $H_X$:

\begin{equation}
\begin{aligned}
H_s & = H_X[\tilde{s}], H_s \in \mathbb{R}^H \\
H_o & = H_X[\tilde{o}], H_o \in \mathbb{R}^H
\end{aligned}
\end{equation}

Finally, the entity pair representation is fused via a fully connected layer with a nonlinear activation to obtain the multimodal entity representation used for relation prediction:

\begin{equation}
H_E = \sigma(W_e [H_s; H_o] + b_e)
\end{equation}

where $H_E \in \mathbb{R}^H$ is entity-pair representation, $\sigma$ is non-linear activation function, and $W_e \in \mathbb{R}^{H \times 2H}$ and $b_e \in \mathbb{R}^H$ are learnable parameters.

\subsection{Relational Semantic Encoder}
\label{sec:rel-encoder}

To enhance the model's ability capturing relational semantics, we introduce natural language descriptions to replace traditional discrete label representations which enables the model to precisely distinguish semantic differences among relation types. 

We first utilize the GPT-4o model to convert each relation label in the training dataset into a natural language description. The generated descriptions are then manually reviewed to ensure both accuracy and semantic consistency.

For each relation description $d_i$, we use an independent BERT encoder (denoted as \texttt{RelEncoder}) to encode the description and obtain a global semantic representation of the relation:

\begin{equation}
\begin{aligned}
X_r & = \text{RelEncoder}(d_i) \\
H_r & = \frac{1}{L_R}\sum_{i=1}^{L_R} X_r[i]
\end{aligned}
\end{equation}

Where $H_r$ denotes the mean-pooled semantic vector representing the relation, which captures the distribution of the relation in the semantic space.

\subsection{Contrastive Semantic Retrieval Strategy}
\label{sec:retrieval}

During model training, the ROC framework abandons traditional classification loss functions and instead constructs a contrastive learning-based semantic retrieval mechanism. Inspired by the SimCLR approach, we optimize the model's ability to discriminate semantic relations by maximizing the cosine similarity between positive samples (i.e., an entity pair and its corresponding relation description) while minimizing the similarity to negative relation descriptions within the same batch. 

The loss function is defined as follows:

\begin{equation}
\mathcal{L} = - \frac{1}{N} \sum_{i=1}^{N} \log \frac{\exp(\text{sim}(H_E^i, H_r^i)/\tau)}{\sum_{j=1}^{N} \exp(\text{sim}(H_E^i, H_r^j)/\tau)}
\end{equation}

where $\text{sim}(\cdot, \cdot)$ denotes cosine similarity, $\tau$ is a temperature hyperparameter, and $H_E^i$ and $H_r^i$ represent the multimodal entity pair representation and the corresponding relation semantic representation for the $i$-th sample, respectively.

\subsection{Inference}

During inference phase, the model calculates the similarity between the multimodal entity pair representation and all relation semantic embeddings, and selects the relation with the highest similarity as the final prediction. This retrieval-based inference approach significantly enhances the interpretability and flexibility of relation extraction, allowing the model to perform dynamic matching based on semantic similarity rather than relying on fixed category labels. Compared to traditional classification methods, this approach demonstrates stronger adaptability and generalization when handling relation semantic shifts or unseen relation types.

\section{Experimental Settings}

\textbf{Dataset} We evaluate our method on two widely used multimodal relation extraction datasets: MNRE\cite{MNRE} and MORE\cite{MORE}, which respectively correspond to social media contexts and object-entity relation extraction tasks. These datasets encompass rich information on image-text alignment and diverse relation types. More details can be found in Appendix~\ref{app:dataset}.

\noindent
\textbf{Evaluation Metrics} We adopt four commonly used evaluation metrics for multimodal relation extraction tasks: Accuracy, Precision, Recall, and F1 score. The F1 score, which balances Precision and Recall, is the main criterion for subsequent performance analysis.

\noindent
\textbf{Baselines} We compare ROC with a range of representative multimodal RE baselines. Early methods include MTB\cite{MTB}, VisualBERT\cite{VisualBERT}, ViLBERT\cite{ViLBERT}, UMT\cite{UMT}, MKGformer\cite{MKGformer}, MEGA\cite{MEGA} and HVPNeT\cite{HVPNeT}. Among more recent advances, IFAformer\cite{IFA} improves visual-textual alignment through prefix networks and early cross-attention. TSVFN\cite{TSVFN} employs a two-stage fusion strategy to mitigate visual noise. PROMU\cite{PROMU} and MOREformer\cite{MORE} enhance relation prediction via prompt-based and object-centric designs. TMR\cite{TMR} leverages diffusion-based generation for robust alignment, while MMIB\cite{MMIB} adopts an information bottleneck to reduce modality noise. VM-HAN\cite{VMHAN} models higher-order relations using multimodal hypergraphs, and CAMRE\cite{CAMRE} introduces LLM-generated image descriptions to improve alignment. APOLLO\cite{APOLLO} proposes a triple contrastive mechanism for cross-modal semantic learning. We also use Qwen-VL\cite{qwen2.5vl}, BLIP2\cite{blip2}, and InstructBLIP\cite{instrcutBLIP} as vision-language LLM baselines to further validate ROC's effectiveness.

\noindent
\textbf{Implementation details} For detailed information on model configuration, training setup, and hyperparameter settings, see Appendix~\ref{app:exp}.

\section{Experimental Results}

\subsection{Main Results}

\begin{table*}[t]
\centering
\small
\setlength{\tabcolsep}{4pt}
\begin{tabularx}{\textwidth}{
  >{\raggedright\arraybackslash}p{4.2cm}
  >{\centering\arraybackslash}X
  >{\centering\arraybackslash}X
  >{\centering\arraybackslash}X
  >{\centering\arraybackslash}X}
\toprule
Model & Accuracy & Precision & Recall & F1-score \\
\midrule
MTB \cite{MTB}        & 75.69 & 64.46 & 57.81 & 60.86 \\
UMT \cite{UMT}        & 77.84 & 62.93 & 63.88 & 63.46 \\
MEGA \cite{MEGA}       & 80.05 & 64.51 & 68.44 & 66.41 \\
HVPNeT \cite{HVPNeT}     & 92.52 & 83.64 & 80.78 & 81.25 \\
IFAformer \cite{IFA}  & 92.38 & 82.59 & 80.78 & 81.67 \\
TSVFN \cite{TSVFN}      & 92.67 & 85.16 & 82.07 & 83.12 \\
MOREformer \cite{MORE} & 82.67 & 82.19 & 82.35 & 82.27 \\
PROMU \cite{PROMU}      & --    & 84.95 & 85.76 & 84.86 \\
TMR \cite{TMR}        & --    & 90.48 & 87.66 & 89.05 \\
MMIB \cite{MMIB}       & --    & 83.49 & 82.97 & 83.23 \\
VM-HAN \cite{VMHAN}     & \underline{92.57} & 85.76 & 84.69 & 85.22 \\
CAMRE \cite{CAMRE}      & \textbf{95.79} & \textbf{91.73} & \underline{90.16} & \underline{90.94} \\
\midrule
ROC (Ours)               & 90.97 (±0.32) & \underline{91.59} (±0.24) & \textbf{90.85} (±0.32) & \textbf{91.22} (±0.23) \\
\bottomrule
\end{tabularx}
\caption{Main experimental results of the ROC model on the MNRE dataset.}
\label{tab:mnre_comparison}
\end{table*}

\begin{table*}[t]
\centering
\small
\setlength{\tabcolsep}{4pt}
\begin{tabularx}{\textwidth}{
  >{\raggedright\arraybackslash}p{4.2cm}
  >{\centering\arraybackslash}X
  >{\centering\arraybackslash}X
  >{\centering\arraybackslash}X
  >{\centering\arraybackslash}X}
\toprule
Model & Accuracy & Precision & Recall & F1-score \\
\midrule
BERT+SG            & 61.79 & 29.61 & 41.27 & 34.48 \\
BERT+SG+Att        & 63.74 & 31.10 & 39.28 & 34.71 \\
\midrule
MEGA \cite{MEGA}        & 65.97 & 33.30 & 38.53 & 35.72 \\
IFAformer \cite{IFA}    & 79.28 & 55.13 & 54.24 & 54.68 \\
MKGformer \cite{MKGformer} & 80.17 & 55.76 & 53.74 & 54.73 \\
VisualBERT \cite{VisualBERT} & 82.84 & 58.18 & 61.22 & 59.66 \\ 
ViLBERT \cite{ViLBERT}  & 83.50 & 62.53 & 59.73 & 61.10 \\
MOREformer \cite{MORE}  & 83.50 & 62.18 & 63.34 & 62.75 \\
VM-HAN \cite{VMHAN}     & 85.57 & 64.76 & 66.69 & 65.71 \\
APOLLO \cite{APOLLO}    & \underline{85.90} & \underline{67.42} & \underline{70.70} & \underline{69.02} \\
\midrule
ROC (Ours)        & \textbf{90.44} (±0.31) & \textbf{68.85} (±1.12) & \textbf{75.40} (±0.57) & \textbf{71.97} (±0.86) \\
\bottomrule
\end{tabularx}
\caption{Main experimental results of the ROC model on the MORE dataset.}
\label{tab:more_comparison}
\end{table*}

\begin{table*}[t]
\centering
\small
\begin{tabular}{lcccccccc}
    \toprule
    \multirow{2.5}{*}{Ablation Setting} & \multicolumn{4}{c}{\textbf{MNRE}}  & \multicolumn{4}{c}{\textbf{MORE}} \\
    \cmidrule(lr){2-5} \cmidrule(lr){6-9}
    & Accuracy  & Precision & Recall & F1  & Accuracy  & Precision & Recall & F1  \\
    \midrule
    ROC (Full Model) & 90.97	 & 91.59	& 90.85	 & 91.22	& 90.44	 & 68.85	 & 75.40	& 71.97 \\
    \hline
    w/o entity encoder & 90.40	& 90.49  & 90.74	& 90.62	  & 87.09	& 62.42	 & 73.94 & 67.69 \\
    $\Delta$ & -0.57 & -1.1 & -0.11 & -0.6 & \underline{-3.35} & \textbf{-6.43} & -1.46 & \underline{-4.28} \\
    
    w/o entity position & 90.15  & 91.00  & 90.12  & 90.56  & 89.10  & 66.22  & 73.32  & 69.59 \\
    $\Delta$ & -0.82	& -0.59	& -0.73	& -0.66	 & -1.34  & -2.63 & -2.08 & -2.38 \\
    
    w/o entity type & 89.16  & 89.57  & 88.96  & 89.26  & 88.37  & 65.60  & 71.82  & 68.57 \\ 
    $\Delta$ & \underline{-1.81}  & \underline{-2.02}  & \underline{-1.89}  & \underline{-1.96}  & -2.07  & -3.25  &  \underline{-3.58}  & -3.4 \\
    
    w/o relation embedding & 88.23	& 89.00  & 87.72	& 88.36	  & 84.85	& 63.88	 & 64.84 & 64.36 \\
    $\Delta$ & \textbf{-2.74} & \textbf{-2.59} & \textbf{-3.13} & \textbf{-2.86} & \textbf{-5.59} & \underline{-4.97} & \textbf{-10.56} & \textbf{-7.61} \\
\hline
\end{tabular}
\caption{Ablation results of the ROC model on MNRE and MORE datasets. Each row removes one component from the full model. $\Delta$ indicates the performance drop compared to the full model.}
\label{tab:ablation_results}
\end{table*}

To comprehensively evaluate the performance of our proposed ROC model on the multimodal relation extraction task, we conducted main experiments on two standard datasets: MNRE and MORE, and compared our method with several representative existing models. The experimental results are shown in Table~\ref{tab:mnre_comparison} and Table~\ref{tab:more_comparison}.

On the MNRE dataset, the ROC model achieved an accuracy of 90.97\%, which is lower than CAMRE (95.79\%). However, it outperformed all other methods in terms of recall (90.85\%) and F1 score (91.22\%). Compared with CAMRE, ROC improved recall by 0.69 percentage points (a relative improvement of 0.77\%) and F1 score by 0.28 percentage points (a relative improvement of 0.31\%). Significance testing on the F1 score shows a 95\% confidence interval of [90.93\%, 91.51\%], indicating that the improvement is statistically significant. Since both models already exceed 90\% on key metrics, even minor improvements demonstrate ROC's stronger ability in identifying positive samples in multimodal relation extraction.

It is worth noting that models such as TMR and CAMRE utilize additional information (e.g., synthetic samples or image descriptions generated by large models) to enhance understanding of image content, thereby improving the accuracy of relation prediction between entities. However, these methods often overlook the modeling of negative samples (i.e., the "None" relation), resulting in limited improvements in recall. The significant recall improvement achieved by ROC indicates that the retrieval-based paradigm incorporating relation semantics allows for a deeper understanding of semantic relations between entities, leading to more accurate predictions than traditional classification approaches relying solely on discrete labels.

The ROC model outperformed existing methods across all evaluation metrics on the more challenging MORE dataset, demonstrating state-of-the-art performance in relation prediction. Specifically, ROC achieved an F1 score of 71.97, outperforming the second-best model APOLLO by 2.95 percentage points (a relative improvement of 4.27\%). In terms of recall, it reached 75.40, an improvement of 4.70 percentage points over APOLLO (a relative improvement of 6.65\%), while accuracy and precision also increased by 5.29\% and 2.12\%, respectively. These results confirm that the retrieval-based paradigm with relation semantics helps the model more comprehensively and accurately predict semantic relations between entities.

Overall, the retrieval-based multimodal relation extraction approach employed by the ROC model effectively aligns entity pairs with their potential semantic relationships. It achieves the best overall F1 scores on both the MNRE and MORE datasets, providing strong evidence of the effectiveness of the ROC model design.

\subsection{Ablation Study}

To evaluate the contribution of each core component in the ROC model to the overall performance, we designed four ablation studies by removing key model modules and observing the resulting performance changes. The experimental results are shown in Table~\ref{tab:ablation_results}.


\begin{itemize}
  \item w/o entity encoder: Removes the Transformer encoder in the multimodal entity pair feature encoding module to assess the effect of cross-modal feature interaction and fusion.
  
  \item w/o entity position: Removes the entity position encoding mechanism to evaluate the impact of positional information on relation prediction. Global average pooling is applied to maintain feature dimensional consistency.
  
  \item w/o entity type: Removes the pre-extracted entity type information to assess the influence of type priors on prediction performance.
  
  \item w/o relation embedding: Replaces the relation semantic encoder with fixed relation label IDs, degrading the model into a conventional classification architecture to evaluate the effectiveness of the semantic retrieval paradigm.
\end{itemize}

According to the results, removing the cross-modal interaction layer (w/o enc) decreased precision by 1.10 and 6.43 percentage points on the MNRE and MORE datasets, respectively. This indicates that the lack of explicit cross-modal feature interaction significantly degrades performance.

Removing the entity position encoding (w/o ent-pos) caused a performance drop across all metrics, with recall declining even more than in w/o enc. This suggests that positional encoding plays a critical role in relation prediction. On one hand, it explicitly marks the positions of the subject and object entities, enhancing the model's ability to distinguish entity pair structures. On the other hand, it constrains the semantic scope for relation modeling, preventing indiscriminate matching based on global text and image features. Without it, the model must rely on global cues to retrieve relations, lowering prediction accuracy and coverage.

When the entity type embedding was removed (w/o ent-type), all metrics on the MNRE dataset declined, with a 2.02 percentage point drop in precision---the second largest drop among all ablation settings. Precision and recall on the MORE dataset decreased by 3.25 and 3.58 percentage points, respectively, with recall experiencing the second-largest decline. These results indicate that entity type information effectively constrains the relational semantic space. Since relation semantics involve explicit meaning and imply subject-object roles and type expectations, entity types help form a constraint-verification mechanism with relation semantics. Without this information, the model struggles to distinguish between semantically similar relations but differ in kind, which harms prediction accuracy.

After removing the relation semantic encoder (w/o relation), the model experienced an average drop of nearly 3 percentage points across all metrics on the MNRE dataset. On the MORE dataset, recall fell by 10.56 percentage points and F1 score by 7.61 points, indicating even more substantial performance degradation. This powerfully demonstrates the critical role of explicit relation semantic modeling in improving multimodal relation extraction accuracy. The relation semantic encoder provides fine-grained semantic constraints, enabling the model to perform relation prediction via semantic matching. Without this module, the model relies solely on fused features for classification, lacking clear semantic guidance. This increases decision uncertainty, especially in cross-modal subject-object scenarios like the MORE dataset.

Overall, the submodules in the ROC model work collaboratively to build a robust cross-modal semantic matching mechanism.

\subsection{Effect of Visual Input on ROC Performance}

\begin{table*}[h]
\centering
\small
\renewcommand{\arraystretch}{1.2} 
\begin{tabular}{lcccccccc}
\hline
\multirow{2}{*}{Method} & \multicolumn{4}{c}{\textbf{MNRE}} & \multicolumn{4}{c}{\textbf{MORE}} \\
\cline{2-9}
                         & Accuracy & Precision & Recall & F1 & Accuracy & Precision & Recall & F1 \\
\hline
Retrieval-based      & 90.97 & 91.59 & 90.85 & 91.22 & 90.44 & 68.85 & 75.40 & 71.97 \\
Classification-based & 74.91        & 76.99        & 75.51        & 76.25        & 74.24        & 45.41        & 51.87        & 48.43        \\
\hline
\end{tabular}
\caption{Comparison of retrieval-based and classification-based methods on MNRE and MORE datasets.}
\label{tab:comparison_classification_and_retrieval}
\end{table*}

\begin{figure}[t]
    \centering
    \includegraphics[width=0.4\textwidth]{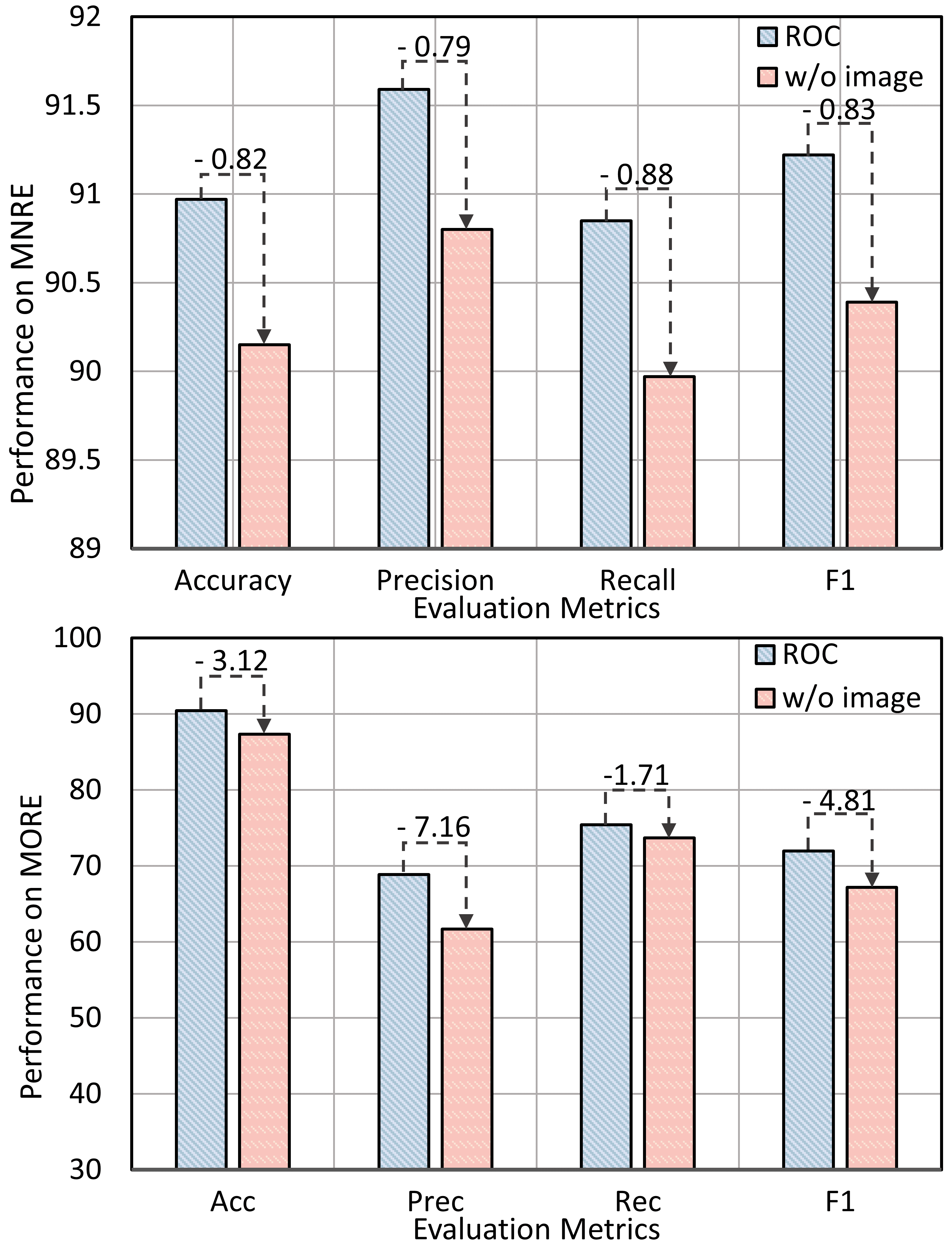}
    \caption{Impact of visual information on ROC performance on MNRE and MORE datasets.}
    \label{fig:fig3}
\end{figure}

To evaluate the impact of visual information on model performance, we conducted a controlled experiment by removing the visual modality. The results in Figure~\ref{fig:fig3} show that the model's performance on the MNRE dataset remains largely unaffected after excluding image inputs. In contrast, a significant performance drop is observed on the MORE dataset. This suggests that the MORE dataset relies more heavily on visual information and further validates the effectiveness of deep multimodal fusion in high-quality vision-language matching tasks.

\subsection{Comparison of Retrieval-based and Classification-based Methods}

To further validate the retrieval-based approach over the classification-based one, we replaced the retrieval module with a standard classification head (a fully connected layer followed by softmax) to compare the performance of both versions. Results are shown in Table~\ref{tab:comparison_classification_and_retrieval}.

As shown in the results, the classification-based version exhibits a significant drop across all metrics. On the MNRE and MORE datasets, the F1 scores decrease by 14.97 and 23.54 percentage points, respectively. This performance gap is primarily because the retrieval-based method leverages semantic matching between the relation descriptions and the input context, while also using entity types to effectively narrow down the candidate space. In contrast, the classification model must select from all possible relation types and cannot fully exploit such contextual information.

\subsection{Comparison with MLLMs}

\begin{table*}[t]
\centering
\small
\begin{tabular}{lcccccccc}
\toprule
\multirow{2}{*}{Model} & \multicolumn{4}{c}{\textbf{MNRE}} & \multicolumn{4}{c}{\textbf{MORE}} \\
\cmidrule(lr){2-5} \cmidrule(lr){6-9}
 & Accuracy & Precision & Recall & F1 & Accuracy & Precision & Recall & F1 \\
\midrule
BLIP2 \cite{blip2}           & 94.86 & 89.42 & 89.84 & 89.63 & 73.09 & 29.91 & 35.54 & 32.48 \\
InstructBLIP \cite{instrcutBLIP}    & 94.98 & 90.11 & 89.69 & 89.90 & 79.48 & 49.33 & 55.24 & 52.12 \\
QwenVL-Plus \cite{qwen2.5vl}     & \textbf{95.57} & 91.41 & 89.84 & 90.62 & -     & -     & -     & -     \\
DeepSeek-V3* \cite{DeepSeek}     & 21.62 & 21.65 & 54.53 & 30.99 & 16.70 & 16.73 & 41.13 & 23.79 \\
\midrule
ROC (Ours)             & 90.97 & \textbf{91.59} & \textbf{90.85} & \textbf{91.22} & \textbf{90.44} & \textbf{68.85} & \textbf{75.40} & \textbf{71.97} \\

\bottomrule
\end{tabular}
\caption{Comparison of the ROC model with MLLMs on the MNRE and MORE datasets.}
\label{tab:comparison_with_mllms}
\end{table*}

To further evaluate the advantages of the ROC model under the current trend of large multimodal language models (MLLMs), we compare it with several representative MLLMs, including fine-tuned versions of BLIP2, InstructBLIP, and Qwen-VL-Plus, as well as the non-fine-tuned DeepSeek-V3. The comparison results are shown in Table~\ref{tab:comparison_with_mllms}.

On the MNRE dataset, the fine-tuned MLLMs achieve high precision, with Qwen-VL-Plus reaching 95.57\%, and InstructBLIP and BLIP2 achieving 94.98\% and 94.86\%, respectively. However, regarding F1 score, the ROC model outperforms all MLLMs, indicating a more balanced performance between precision and recall. Although MLLMs perform competitively in some metrics, they still lag behind ROC in recall and overall stability.

On the more challenging MORE dataset, MLLMs' accuracy remains relatively high, but both precision and recall drop significantly, leading to much lower F1 scores compared to ROC. This performance gap may be attributed to increased semantic complexity, which makes MLLMs more prone to overfitting on the training set and less capable of generalizing to diverse samples.

DeepSeek-V3, as non-fine-tuned MLLM, only performs zero-shot inference in experiments. Its performance is significantly worse than the fine-tuned models and ROC, suggesting that current MLLMs struggle to handle structured extraction tasks without task-specific adaptation.

While MLLMs can achieve competitive results under sufficient resources and tuning conditions, their training costs and adaptation thresholds are relatively high. In contrast, with its lightweight design, the ROC model achieves the best overall performance on both datasets and consistently leads in key metrics such as F1 score, demonstrating superior practicality and deployability.

Additional analyses, including prompt templates, encoder architecture variations, and attention distributions, are provided in Appendix~\ref{sec:visual_encoder}–\ref{sec:attention_weight}.

\section{Conclusion}

We introduce a relation-semantic retrieval-based method for multimodal relation extraction, named ROC. It integrates entity-centric multimodal encoding, position-aware structural modeling, and relation-aware semantic retrieval, showing robust performance across diverse scenarios. Experiments demonstrate that ROC outperforms baselines on the MNRE and MORE datasets, including fine-tuned large-scale MLLMs, particularly excelling in F1 scores. Ablation studies further confirm the contributions of each key component, highlighting the critical roles of explicit multimodal interaction and structured semantic modeling. ROC overcomes the limitations of traditional classification methods in label semantic representation and fine-grained semantic differentiation, offering a novel paradigm for multimodal relation extraction.

\section*{Limitations}

The limitations of our approach are as follows: Our experiments show that zero-shot multimodal LLMs cannot directly perform multimodal relation extraction. However, we have not yet systematically verified whether fine-tuned LLMs can significantly enhance task performance, especially when class labels are reformulated as natural language descriptions. Furthermore, although replacing discrete class labels with semantic descriptions is theoretically applicable to a wide range of classification tasks, the generalizability of this method has not been thoroughly evaluated across diverse domains, tasks, and datasets.

\section*{Ethics Statements}

Our model infers potential relations between entities from text and images, but these are based solely on input content and do not reflect verified real-world facts. The datasets may include personal information and perform basic checks for identifiable or offensive content, but named entities central to the task cannot be anonymized. No human annotators or evaluators are involved; all experiments and evaluations are automated. GPT-4o generates the relation descriptions. The AI tools are used only for grammar correction and relation description generation.

\section*{Acknowledgements}

This work is funded by the National Natural Science Foundation of China (No.62576085), Zhiyuan Laboratory (NO. ZYL2024021) and CCF-Zhipu AL Large Model Fund (NO.202221).

\bibliography{custom}

\appendix

\section{Licenses}

The models used in this work, including ViT-B/16 \cite{ViT} and BERT \cite{BERT}, are licensed under the Apache License 2.0. The GPT series models are developed and released by OpenAI under their respective terms of use. Detailed license information is available on the official GitHub repositories or documentation pages.

We use the Stanford Named Entity Recognizer (NER) tool for entity recognition, which is distributed under the GNU General Public License v2 (GPLv2).\footnote{\url{https://nlp.stanford.edu/software/CRF-NER.html}}

The datasets used include the MNRE dataset\cite{MNRE}, with details available on its GitHub page, and the MORE dataset\cite{MORE}, which is released under the MIT License.

In summary, all licenses permit academic research use.

\begin{table*}[t]
    \centering
    \small
    \begin{tabular}{lcccccc}
    \toprule
    Dataset & \#Image & \#Word & \#Sentence & \#Entity & \#Relation & \#Instance \\
    \midrule
    MNRE    & 9,201   & 258k   & 9,201      & 30,970   & 23         & 15,485  \\
    MORE    & 3,559   & -      & 3,559      & -        & 21         & 3,559 \\
    \bottomrule
    \end{tabular}
    \caption{Detailed information on MNRE and MORE datasets.}
    \label{tab:dataset_info}
    \end{table*}

\section{Dataset Details}
\label{app:dataset}
In the context of increasingly rich multimedia data, extracting structured information from multimodal data that includes both text and images has become particularly important. To promote research in this field, the academic community has introduced specially designed datasets to support the development of relation extraction tasks.
\textbf{MNRE} \cite{MNRE} is a specially designed dataset aimed at evaluating and enhancing the capabilities of neural relation extraction models, with a particular emphasis on the importance of incorporating visual evidence in social media posts. The dataset contains over 9,000 sentences covering 23 distinct relation types, sourced from Twitter and annotated by crowd-sourced workers. Each sentence is paired with a relevant image, intended to supplement contextual information that may be missing from the text alone, thereby aiding in the more accurate identification of relationships between entities.
\textbf{MORE} \cite{MORE} is a novel dataset focused on extracting object-entity relations from both text and images, developed by a research team from Nanjing University. It consists of 3,559 pairs of news headlines and their corresponding images, annotated with 20,264 multimodal relational facts across 21 relation types, involving 13,520 visual objects with an average of 3.8 objects per image. MORE is designed to pose challenges to existing methods in handling complex relationships between text and images, particularly emphasizing scenarios that require identifying relations between entities and visual objects across different modalities. This dataset serves as an important resource for advancing research on multimodal relation extraction. With MORE, researchers can explore how to enhance models' ability to understand the interactions between textual and visual information.

\section{Implementation Details}
\label{app:exp}
To ensure the fairness and rigor of our conclusions, we adopted the same text encoder as used in previous methods. We conducted comprehensive experiments on both the MNRE and MORE datasets. Specifically, the model was trained for 50 epochs with a batch size of 32, using the AdamW optimizer and a hidden layer dimension 768. The overall model contains approximately 342.52 million parameters, including the BERT-base-uncased text encoder (109.48 million parameters), the ViT-base-patch32-384 image encoder (88.12 million parameters), and the BERT-base-uncased relation encoder (109.48 million parameters). To ensure reproducibility, all experiments were conducted under the following setup: CPU was Intel(R) Xeon(R) CPU E5-2620 v4 @ 2.10GHz with 32 cores, memory size was 128GB, and GPU was NVIDIA RTX 8000 with 48GB VRAM. The operating system was Ubuntu 16.04.7 LTS, with CUDA version 11.7, PyTorch version 1.13.1, and Python version 3.10.4.

The experimental results are obtained by one-shot inference with a random seed of 648, and the results are reproducible.

\section{Prompt Template for Extending Natural Language Descriptions}
\label{sec:template}

This study employs GPT-4o to generate explanations and descriptions for the predefined set of relations in the MNRE and MORE datasets. Table~\ref{tab:relation_prompt} presents the specific prompt and several example relation descriptions generated by large model.

\begin{table*}[t]
\centering
\small
\begin{tabularx}{\textwidth}{|X|}
\hline
\textbf{Relation Description Few-shot Prompt Template} \\
\hline
Given a set of relation labels, convert them into a JSON format where each key is a relation label and the value is a clear, textual explanation of the relationship that the label represents. If the provided labels include numeric indices, please remove them in your output.

There are some relation description examples: \\
\{ \\
\quad "None": "Indicates that there is no relationship between the subject and object entity, based on text and image information. Subject and object can be of any type.", \\
\quad "/org/loc/locate\_at": "Indicates that an organization is located at a specific geographical location (e.g., a company headquartered in a city)", \\
\quad "/per/misc/awarded": "Specifies an award, honor, or recognition received by a person (e.g., a Nobel Prize or medal)." \\
\} \\
\hline
\textbf{Example of Generated Relation Descriptions and Label Set} \\
\hline
\{ \\
\quad "None": "Indicates that there is no relationship between the subject and object entity, based on text and image information. Subject and object can be of any type.", \\
\quad "/loc/loc/contain": "Describes a containment relationship where both the subject and object are geographical locations, with the subject containing the object (e.g., a country containing a city).", \\
\quad "/misc/loc/held\_on": "Indicates that a miscellaneous event or activity occurred at a specific geographical location (e.g., a festival held in a city).", \\
\quad "/misc/misc/part\_of": "Describes a part-whole relationship between two miscellaneous entities (e.g., a chapter being part of a book).", \\
\quad "/org/loc/locate\_at": "Indicates that an organization is located at a specific geographical location (e.g., a company headquartered in a city).", \\
\quad "/org/org/alternate\_names": "Lists alternative names, aliases, or abbreviations for an organization (e.g., a company and its trade name).", \\
\quad "/org/org/subsidiary": "Describes a parent-subsidiary relationship between two organizations, with the subject being the parent and the object being the subsidiary (e.g., a corporation and its owned companies).", \\
\quad "/per/loc/place\_of\_birth": "Indicates the geographical location where a person was born (e.g., a city or country)." \\
\} \\
\hline
\end{tabularx}
\caption{Prompt template and generated relation descriptions for few-shot relation label explanation.}
\label{tab:relation_prompt}
\end{table*}

\section{Impact of Visual Encoder Architectures on Model Performance}
\label{sec:visual_encoder}

\begin{figure}[t]
    \centering
    \includegraphics[width=0.4\textwidth]{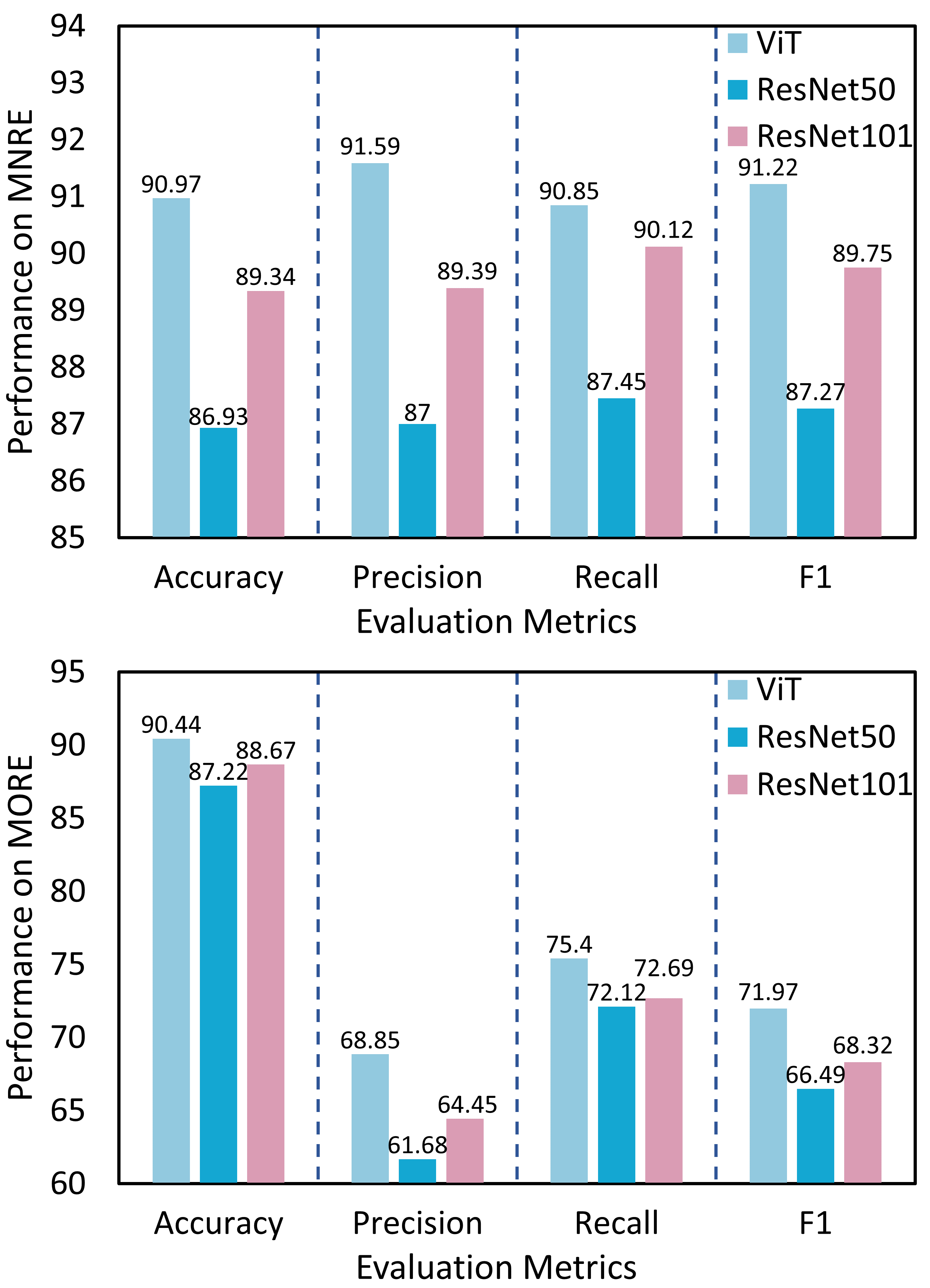}
    \caption{Effect of visual encoder on relation prediction performance on MNRE and MORE datasets.}
    \label{fig:fig4}
\end{figure}

To evaluate the performance of different visual encoders in multimodal relation extraction tasks, we replaced the visual feature extraction module in the ROC model with ResNet50, ResNet101 (CNN-based architectures), and ViT-B/16 (a Transformer-based architecture). We conducted comparative experiments on the MNRE and MORE datasets, and the results are shown in Figure~\ref{fig:fig4}.

On the MNRE dataset, ViT outperforms the ResNet-based encoders across all metrics. Specifically, its F1 score is 3.95 percentage points higher than that of ResNet50 and 1.47 points higher than that of ResNet101, indicating stronger stability and generalization capability. On the more complex MORE dataset, which has a higher dependency on visual semantics, ViT achieves even greater improvements, with F1 scores 5.48 and 3.65 percentage points higher than ResNet50 and ResNet101, respectively. The precision gain is particularly notable, with a 7.17-point increase over ResNet50.

The primary reason for this performance difference lies in the architectural consistency and its impact on modality fusion. In the ROC model, the text encoder and the relation semantics encoder are based on the Transformer architecture. ViT, as a structurally homogeneous visual encoder, adopts a similar approach to feature extraction and semantic modeling as BERT, utilizing self-attention mechanisms to model global dependencies. This architectural compatibility naturally facilitates more efficient semantic alignment in the fusion stage.

In contrast, as CNN-based encoders, the ResNet series emphasize local feature extraction. Their localized modeling mechanisms are structurally heterogeneous to the global modeling strategy of Transformers, introducing additional alignment challenges during fusion and thus reducing the quality of modality integration. Since multimodal relation extraction tasks rely heavily on global semantic reasoning between text and image, Transformer-based encoders are inherently more suitable.

\section{Impact of Encoder Depth on Model Performance}
\label{sec:encoder_depth}

\begin{figure}[t]
    \centering
    \includegraphics[width=0.4\textwidth]{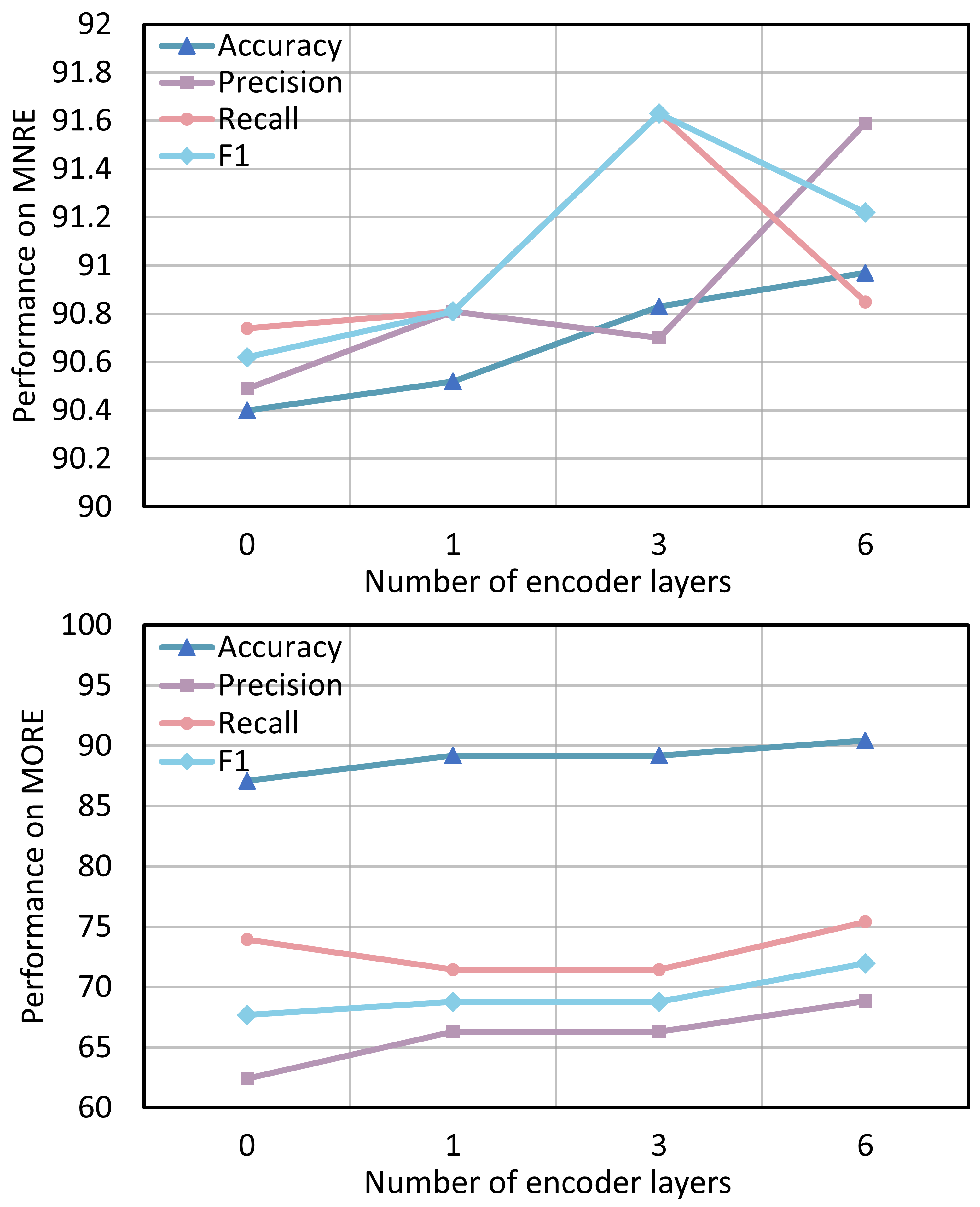}
    \caption{Impact of different encoder numbers on modal fusion performance on MNRE and MORE datasets.}
    \label{fig:fig5}
\end{figure}

\begin{table*}[t]
\centering
\renewcommand{\arraystretch}{1}
\small
\begin{tabular}{l c l c}
\toprule
\multicolumn{2}{c}{\textbf{SciRE}} & \multicolumn{2}{c}{\textbf{NYT}} \\
\cmidrule(lr){1-2} \cmidrule(lr){3-4}
Model & F1 & Model & F1 \\
\midrule
MTB \cite{MTB}         & 87.4  & BiRTE \cite{BiRTE}      & 92.8 \\
REBEL \cite{REBEL}      & 87.7  & DirectRel \cite{DirectRel}   & 92.9 \\
IRE-RoBERTa \cite{IRE}         & 88.9  & GRTE \cite{GRTE}        & 93.4 \\
RELA \cite{RELA}        & 90.3  & OD-RTE \cite{OD-RTE}      & 93.9 \\
ROC (ours)  & 88.39 & ROC (ours) & 91.36 \\
\bottomrule
\end{tabular}
\caption{Performance comparison on the SciRE and NYT datasets. SciRE results are from \cite{RELA}, and NYT results are from \cite{OD-RTE}}
\label{tab:scire_nyt}
\end{table*}

As a key component of modality fusion, the Transformer encoder in the ROC model models the attention-based interactions between visual and textual information. This section investigates the impact of different encoder depth on model performance through comparative experiments, with the results shown in Figure~\ref{fig:fig5}.

When no encoder is introduced on the MNRE dataset, the model achieves an accuracy of 90.40\% and an F1 score of 90.62\%. With the addition of a single encoder layer, these metrics increase to 90.52\% and 90.81\%, respectively, indicating that the attention mechanism contributes positively to basic multimodal fusion. The model performance improves as the encoder depth increases to 3 and 6 layers. However, the F1 score of the 3-layer configuration slightly surpasses that of the 6-layer one, with an improvement of 0.41 percentage points. This difference is primarily due to a slight drop in recall for the 6-layer structure, suggesting that while deeper encoders enhance prediction precision, they may also suppress the recognition of marginal samples, thereby affecting recall.

On the MORE dataset, the model shows a more significant response to increasing encoder depth. Without any encoder, the model achieves an accuracy of 87.09\% and an F1 score of 67.69\%. When the encoder depth is increased to 6 layers, the F1 score improves by 4.55 percentage points, with all metrics reaching their highest values. This indicates that deep attention mechanisms play a substantial role in enhancing cross-modal semantic fusion on this dataset.

The performance differences between the two datasets can be attributed mainly to the varying quality of image-text alignment. The MNRE dataset suffers from relatively weak semantic associations between images and text, where shallow fusion helps preserve more original semantics and allows the model to learn alignment strategies autonomously. In contrast, the MORE dataset is constructed from news articles with well-aligned image-text pairs, where deeper multimodal interaction more effectively captures applicable semantics. Therefore, deeper encoder layers are more beneficial for modeling complex semantic relationships in multimodal scenarios with structurally complete and semantically consistent inputs.

In summary, the optimal depth of the encoder should be adapted to the dataset's characteristics: shallow structures are better suited for tasks with loose image-text alignment, while deeper structures are more effective in scenarios with strong semantic coupling between modalities, promoting semantic aggregation and relation understanding.

\section{Experiments on Text-Only Datasets}
\label{sec:text_only}

We conducted experiments on two text-only datasets from different domains: SciRE, which focuses on scientific papers, and NYT, which covers the news domain, in order to evaluate the applicability of ROC across diverse professional text datasets. The results are presented in Table~\ref{tab:scire_nyt}.

As shown in the results, on the SciRE dataset, ROC performs only 1.54 percentage points lower than RELA, while outperforming MTB and REBEL. On the NYT dataset, ROC is 2.54 percentage points lower than OD-RTE. These findings demonstrate that ROC remains competitive and stable across different domain-specific text datasets.

\section{Case Study}
\label{sec:case_study}

To evaluate the ability of the ROC model to distinguish semantically similar relations, we selected a subset of closely related relations from the MNRE and MORE datasets and constructed relation pairs. In the experiments, the model was required to identify the correct relation within each pair, effectively performing a binary classification task. The experimental results are shown in Table~\ref{tab:relation_pairs}.

\begin{table}[t]
\centering
\renewcommand{\arraystretch}{1.2}
\setlength{\tabcolsep}{0pt}
\small
\begin{tabular}{l c}
\hline
\textbf{MNRE} & \textbf{Accuracy} \\
\hline
/per/misc/nationality vs /per/misc/race & 100.00 \\
/per/per/peer vs /per/per/neighbor & 94.94 \\
/org/loc/locate\_at vs /loc/loc/contain & 92.41 \\
/per/misc/present\_in vs /per/loc/place\_of\_residence & 90.29 \\
/per/org/member\_of vs /per/per/alumi & 90.00 \\
\hline
\textbf{MORE} & \textbf{Accuracy} \\
\hline
/org/loc/locate\_at vs /org/misc/present\_in & 100.00 \\
/per/misc/president vs /per/org/leader\_of & 97.73 \\
/per/misc/nationality vs /per/misc/party & 98.97 \\
/per/per/relatives vs /per/per/partner & 92.65 \\
/per/misc/present\_in vs /org/misc/present\_in & 91.74 \\
\hline
\end{tabular}
\caption{Accuracy of ROC on semantically related relation pairs from the MNRE and MORE datasets.}
\label{tab:relation_pairs}
\end{table}

As shown in the results, ROC attains consistently high accuracy across most relation pairs. For pairs with clear semantic distinctions, such as \texttt{/per/misc/nationality\allowbreak-/per/misc/race} and \texttt{/org/loc/locate\_at\allowbreak-/org/misc/present\_in}, the model achieves 100\% accuracy. For pairs that are semantically similar but differ in entity types, such as \texttt{/per/misc/present\_in\allowbreak-/per/loc/place\_of\_residence} and \texttt{/per\allowbreak/misc/present\_in-/org/misc/present\_in}, the accuracies are 90.29\% and 91.74\%, respectively, demonstrating that ROC can effectively exploit entity type information to distinguish subtle relation differences.

\section{Visualization of Attention Weight Distribution}
\label{sec:attention_weight}

\begin{figure*}[t]
    \centering
    \includegraphics[width=\textwidth]{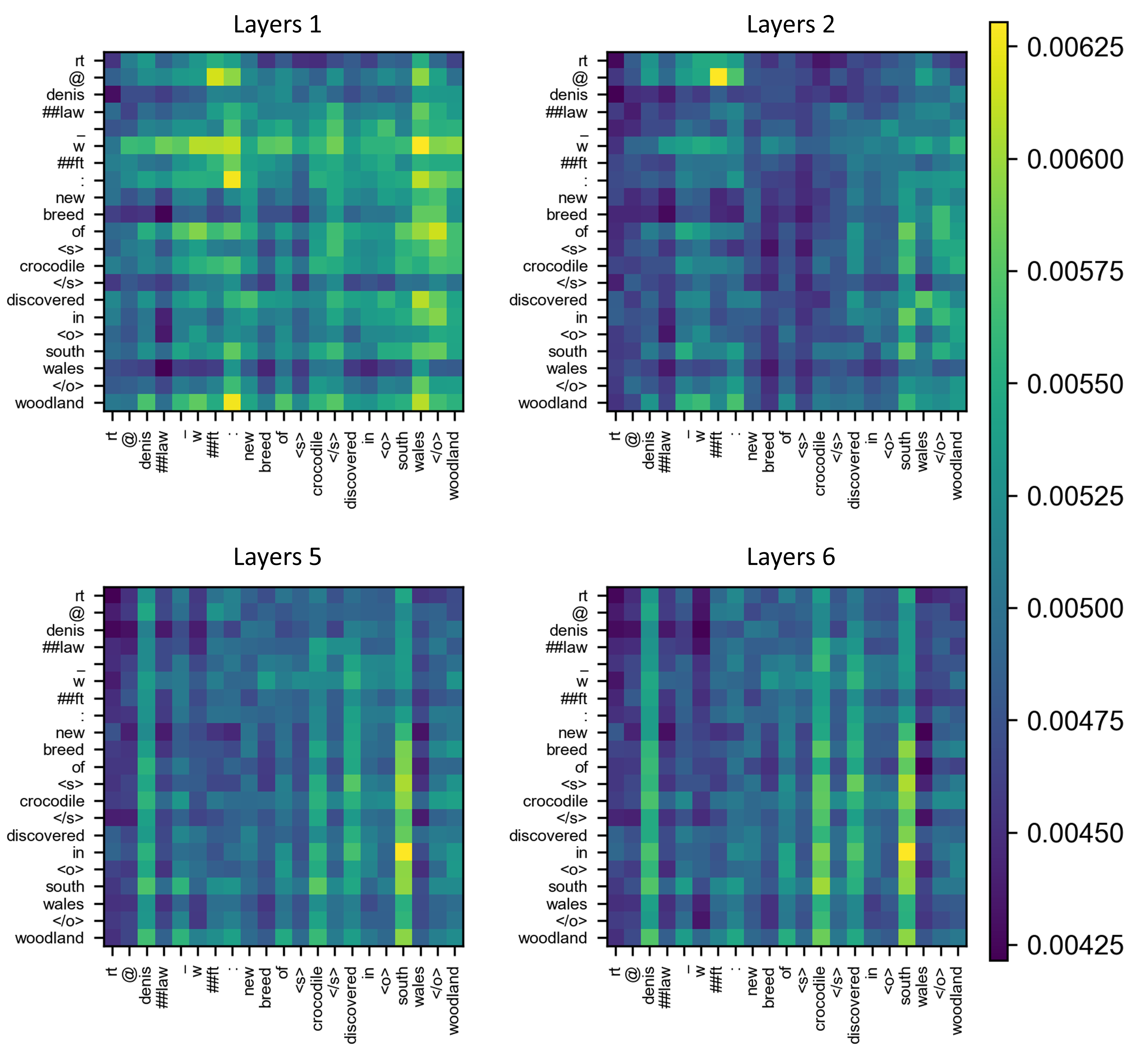}
    \caption{Visualization of attention weight distribution.}
    \label{fig:attention}
\end{figure*}

To further validate the ROC model's semantic modeling capability and interpretability in processing specific samples, we conducted a visualization analysis of the attention distribution in the feature encoder for multimodal entity pairs. As shown in Figure~\ref{fig:attention}, a real-world example was selected: \textit{"RT @DenisLlaw\_WFT: New breed of Crocodile discovered in South Wales woodland",} where the subject is "Crocodile", the object is "South Wales", and the relation type is \texttt{/misc/loc/held\_on}.

The figure illustrates the attention weight distribution of query vectors across different encoding layers. In layers 1 and 2, the attention distribution is relatively dispersed and does not focus on key entities, indicating that at this stage, the model mainly captures global semantic features without explicitly concentrating on the entity pair. In contrast, layers 5 and 6 show a progressively focused attention pattern, with most attention weights concentrated on the words "crocodile" and "south", which correspond to the start positions of the entity pair. This demonstrates the model's dynamic transition from integrating global semantics to identifying local entities layer by layer.

Moreover, in layers 5 and 6, the query word "in" exhibits high attention towards the object word "south", consistent with the semantic alignment of "loc" in the predicted relation \texttt{/misc/loc/held\_on}, indicating that the model has captured semantic cues representing spatial location. By contrast, "in" shows lower attention towards the subject entity, aligning with the weaker type constraint of the "misc" label for the subject in this relation type.

Overall, the ROC model achieves hierarchical semantic modeling through its multi-layer encoder: shallow layers focus on context and global information, while deeper layers progressively concentrate on key entities and capture latent relational semantics. Additionally, the attention mechanism effectively filters redundant information, reducing interference from irrelevant words during training, thereby enhancing both the interpretability and robustness of the model.

\end{document}